\documentclass[letterpaper]{article} 
\usepackage{aaai25}  
\usepackage{times}  
\usepackage{helvet}  
\usepackage{courier}  
\usepackage[hyphens]{url}  
\usepackage{graphicx} 
\usepackage{natbib}  
\usepackage{caption} 
\usepackage{algorithm}
\usepackage{algorithmic}
\usepackage{amsmath}
\usepackage{amssymb} 

\usepackage{graphicx}
\usepackage{booktabs}

\usepackage{newfloat}
\usepackage{listings}
\DeclareCaptionStyle{ruled}{labelfont=normalfont,labelsep=colon,strut=off} 
\floatstyle{ruled}
\newfloat{listing}{tb}{lst}{}
\floatname{listing}{Listing}
\title{MindMem: Multimodal for Predicting Advertisement
Memorability Using LLMs and Deep Learning}
\author {
    Sepehr Asgarian\textsuperscript{\rm 1},  
    Qayam Jetha\textsuperscript{\rm 1},  
    Jouhyun Jeon\textsuperscript{\rm 1}
}
\affiliations {
    \textsuperscript{\rm 1}Klick Health \\
    \{sasgarian, qjetha, cjeon\}@klick.com
}

\usepackage{bibentry}

\begin{document}

\maketitle

\begin{abstract}
In the competitive landscape of advertising, success hinges on effectively navigating and leveraging complex interactions among consumers, advertisers, and advertisement platforms. These multifaceted interactions compel advertisers to optimize strategies for modeling consumer behavior, enhancing brand recall, and tailoring advertisement content. To address these challenges, we present MindMem, a multimodal predictive model for advertisement memorability. By integrating textual, visual, and auditory data, MindMem achieves state-of-the-art performance, with a Spearman’s correlation coefficient of 0.631 on the LAMBDA and 0.731 on the Memento10K dataset, consistently surpassing existing methods. Furthermore, our analysis identified key factors influencing advertisement memorability, such as video pacing, scene complexity, and emotional resonance. Expanding on this, we introduced MindMem-ReAd (MindMem-Driven Re-generated Advertisement), which employs Large Language Model-based simulations to optimize advertisement content and placement, resulting in up to a 74.12\% improvement in advertisement memorability. Our results highlight the transformative potential of Artificial Intelligence in advertising, offering advertisers a robust tool to drive engagement, enhance competitiveness, and maximize impact in a rapidly evolving market. 
\end{abstract}

\section{Introduction}
 The advertising industry operates within a highly competitive landscape, where the ability to capture and sustain consumer attention is paramount. The intricate interactions among consumers, advertisers, and platforms within multi-agent strategic settings are crucial for businesses to effectively navigate this complex landscape. These settings enable the simulation of diverse consumer behaviors, brand recall and engagement, and platform optimizations, allowing advertisers to refine their strategies—from understanding consumer preferences to fine-tuning advertisement placements and crafting persuasive messages. Predicting advertisement memorability is crucial to bridge the gap between understanding consumer interactions and crafting advertisements that effectively capture and retain consumer attention.

Deep learning algorithms and large language models (LLMs) have significantly improved our ability to predict and enhance advertisement memorability. Memorability is a critical driver of consumer engagement, brand loyalty, and purchase decisions, yet it remains a challenging factor to measure and optimize due to its inherent complexity. By integrating textual, visual, and auditory data, Deep Learnings and LLMs provide a more comprehensive understanding of the elements that amplify an advertisement's impact \cite{li2022adaptive}. However, many existing methods are limited by their reliance on single-modal data and their inability to account for the complexities of human cognition. To effectively model human cognition and memorability, a multimodal approach is essential, as it more closely mirrors the way humans perceive and process information from their environment \cite{wang2024comprehensive}. By leveraging such multimodal datasets and developing adaptive multimodal ensemble methods, advertisers are allowed to craft impactful content but also simulate long-term consumer engagement within multi-agent strategic setting.

In this study, we introduce MindMem, a multimodal framework for predicting advertisement memorability, and MindMem-ReAd (MindMem-Driven Re-generated Advertisement), a scalable method for enhancing memorability by fine-tuning language models on advertisement datasets. Focusing on the advertiser's role within multi-agent strategic settings, these tools demonstrate how generative AI can bridge the gap between theoretical multi-agent strategies and practical advertising solutions. Our approach aims to assist businesses in creating more targeted, memorable, and effective campaigns in an incresingly competitive market.

\section{Related Work}
\subsection{Factors influencing Memorability}
Bainbridge et al. explored how humans process and retain visual stimuli, emphasizing the importance of emotionally salient and visually distinctive elements in enhancing memorability \cite{khosla2013modifying}. Their findings suggested that humans are more likely to remember visual scenes that contain unique or emotionally charged content, as opposed to mundane or repetitive scenes. Additionally, the other study focused on the concept of intrinsic memorability revealing that certain visual characteristics, such as color, object saliency, and scene composition, naturally influence memory retention, independent of individual viewer biases \cite{khosla2015understanding}. These studies laid the groundwork for understanding the cognitive processes involved in memorizing visual information and provided key insights into the types of visual content that are more likely to be remembered.
Several studies have aimed to identify the specific features or characteristics that contribute to the memorability of visual content. One study assessed the memorability of various objects within scenes, and found that certain object categories, like faces and animals, are inherently more memorable than others, such as buildings or landscapes \cite{isola2011understanding}. It highlighted the role of object prominence and scene context in shaping human memory. Similarly, it has been shown that the memorability of a scene is largely driven by its most memorable object \cite{7410487}. Despite these valuable insights, these studies were limited in their focus on static images and often failed to account for the dynamic, multimodal nature of real-world stimuli, such as advertisements or videos. Moreover, these works largely overlooked the temporal and emotional dimensions that play a critical role in memory formation.

\subsection{Machine Learning Approaches for Multimodal Memorability Prediction}
More recently, multimodal approaches have emerged as a powerful method for memorability prediction, integrating visual, textual, and audio features to capture a broader spectrum of the factors that contribute to memory retention. Several studies have investigated predicting memorability from video content, integrating audio and emotional cues to enhance model accuracy \cite{dudzik2020investigating}. Other study leveraged video-triggered Electroencephalogram (EEG) data to examine how emotions evoked by videos influence memorability \cite{hu2020video}. Another study has integrated LLMs with deep learning to process not just visual features, but also audio and textual elements, highlighting the benefit of capturing the complex interactions across modalities in advertisements \cite{harinilong}. Although these models have improved prediction accuracy, they often fail to fully capture the complexity of human cognition, as they process modalities separately rather than integrating them into cohesive multimodal representations: an essential aspect for modeling human memory, particularly in scenarios requiring temporal processing and adaptability.

\section{Methods}
\subsection{Dataset}
To develop and evaluate MindMem, we use two datasets, Long-term Ad MemoraBility DAtaset (LAMBDA) \cite{harinilong} and Memento10K \cite{newman2020multimodal}, which provide complementary settings for assessing advertisement memorability and general video memorability.

To train and build our models to predict advertisement memorability, we used the LAMBDA dataset. The dataset consists of 2,205 commercial advertisements from 276 brands across 113 industries. The LAMBDA dataset includes videos released between 2008 and 2023, with an average duration of 33 seconds. These videos feature diverse characteristics, such as varying scene velocities, the presence of humans or animals, visual and audio branding, emotional content, scene complexity, and different audio types. Participants viewed those advertisements, and their brand recall, advertisement recall, scene recall, and audio recall were assessed after a minimum of 24 hours. Memorability scores were calculated by averaging brand recall scores from 1,749 participants to determine the overall long-term advertisement memorability. The memorability scores were scaled ranging from 0 to 1. In total, 1,963 advertisements with memorability scores were used to train models, and 219 used to test model performance. Percentage of speech in a video, video length, and time of day to watch advertisements showed non-significant correlations with memorability score. Meanwhile, negative emotions are more memorable than positive emotions \cite{harinilong}. Video popularity and memorability show a positive correlation.

To assess the reliability of the MindMem architecture, we evaluate it using the Memento10K dataset \cite{newman2020multimodal}. This dataset was constructed by scraping natural videos from the Internet and filtering out artificial scenes and undesirable features (e.g., watermarks), resulting in a collection of 10,000 videos. The dataset emphasizes both the visual and semantic aspects of video memorability and includes human-annotated memorability scores, action labels, and textual descriptions (five human-generated captions per video). It is partitioned into training (7,000 videos), validation (1,500 videos), and test (1,500 videos) sets. For our analysis, we applied the MindMem architecture to the training set and evaluated its performance on the validation set, the results of which are presented here.

\subsection{Multimodal Data Embeddings}
In MindMem, we leverage pre-trained LLMs as our cognitive modules (Figure~\ref{fig:mindmem}). For video embedding, Long Video Assistant (LongVA) model was used to extract visual features from the dataset \cite{zhang2024long}. By leveraging the last hidden layer of the LongVA, we capture both visual and temporal information from long video sequences. For audio embedding, we first extracted audio from videos and fed them into Qwen2 (7B) audio model \cite{Qwen2-Audio}, leveraging its last hidden layer to produce audio embeddings. For text embedding, Gemini Pro 1.5 \cite{team2024gemini} was used to generate detailed textual descriptions of video content by posing targeted questions about scenes and visual details (Appendix 1). These descriptions were then processed by the Qwen2 (7B) text model \cite{yang2024qwen2}, which extracted embeddings from the last hidden layer. 

\subsection{Model Generation and Evaluation}

\begin{figure}[t]
    \centering
    \includegraphics[height=7cm]{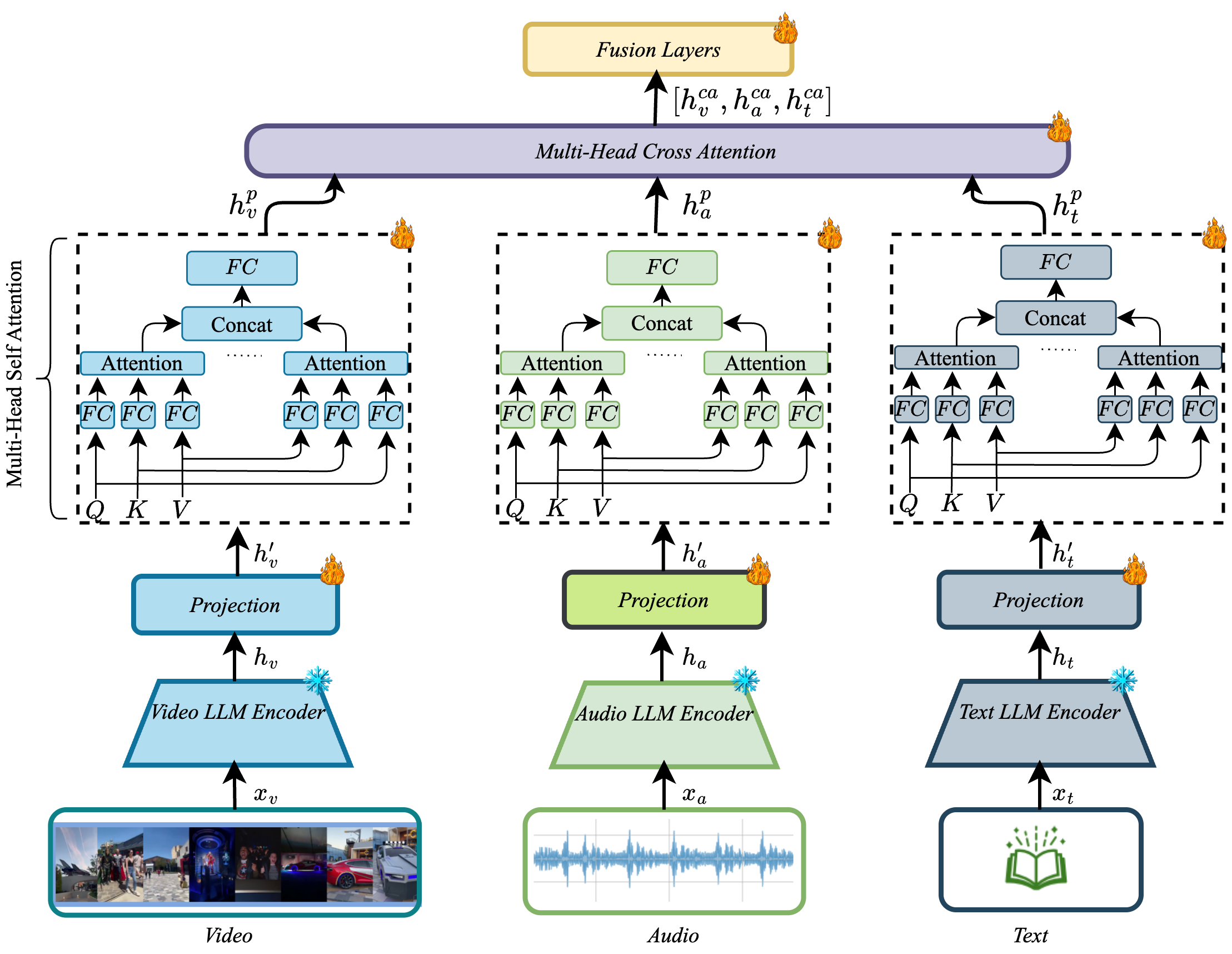} 
    \caption{Architecture of the MindMem model for predicting advertisement memorability. The model processes visual, auditory, and textual inputs using pre-trained embedding models (indicated by snowflake icons), such as encoder LLMs for audio, video, and text, which remain non-trainable (frozen) during training. These embeddings are then fed into trainable components (indicated by fire icons). The trainable layers include projection layers that align embeddings into a shared latent space, multi-head self-attention layers that capture intra-modal dependencies, and multi-head cross-attention layers that integrate information across modalities. Finally, a fusion layer combines the attended embeddings to predict the memorability score.}
    \label{fig:mindmem}
\end{figure}

Figure~\ref{fig:mindmem} shows the procedure to train and build MindMem. As described previously, visual, auditory, and textual embeddings are performed, and the encoded representations of those modalities are expressed:

\begin{equation}
\begin{aligned}
h_v &= \text{LongVA}(x_v), \\
h_a &= \text{Qwen2\_Audio}(x_a), \\
h_t &= \text{Qwen2\_Text}(x_t),
\end{aligned}
\end{equation}

where $x_v$, $x_a$, and $x_t$ are the raw inputs for visual, auditory, and textual data, respectively, while $h_v$, $h_a$, and $h_t$ represent their corresponding embeddings.

To predict memorability scores, the MindMem architecture processes these embeddings through several key components, which are detailed below:

\subsubsection{Projection Layers}

To ensure compatibility across modalities, the extracted embeddings ($\mathbf{h}_v$, $\mathbf{h}_a$, $\mathbf{h}_t$) are projected into a shared latent space of dimension 1,024. This involves linear transformation, layer normalization, and dropout:
\begin{equation}
\begin{aligned}
\mathbf{h}_v' &= \text{Dropout}(\text{LayerNorm}(\text{Linear}(\mathbf{h}_v))), \\
\mathbf{h}_a' &= \text{Dropout}(\text{LayerNorm}(\text{Linear}(\mathbf{h}_a))), \\
\mathbf{h}_t' &= \text{Dropout}(\text{LayerNorm}(\text{Linear}(\mathbf{h}_t))).
\end{aligned}
\end{equation}

where, $\mathbf{h}_v$, $\mathbf{h}_a$, and $\mathbf{h}_t$ represent the initial embeddings from the visual, auditory, and textual modalities, while $\mathbf{h}_v'$, $\mathbf{h}_a'$, and $\mathbf{h}_t'$ are the projected embeddings. These transformations reduce the original dimensionality while preserving the essential features necessary for downstream tasks.

\subsubsection{Self-Attention Pooling
}

Since visual, audio, and text embeddings have variable sequence lengths, we use self-attention pooling to aggregate each modality’s embeddings into fixed-length vectors. This process captures intra-modal dependencies and emphasizes the most relevant features. 

 Self-attention operates on the query $Q$, key $K$, and value $V$ vectors, which are derived from the modality embeddings $h_v'$, $h_a'$ and $h_t'$. The formula for self-attention is as follows:

\begin{equation}
\text{Attn}(Q, K, V) = \text{softmax}\left( \frac{QK^\top}{\sqrt{d_k}} \right)V
\end{equation}

where $d_k$ is the dimensionality of the query and key vectors. Notably, in the self-attention mechanism, $Q$, $K$, and $V$ are representations of the same modality.

These self-attention outputs are then pooled to produce fixed-length representations for each modality. This yields the fixed-length pooled representations $\mathbf{h}_v^p$, $\mathbf{h}_a^p$, and $\mathbf{h}_t^p$ for the visual, auditory, and textual modalities, respectively.

\begin{equation}
\begin{aligned}
\mathbf{h}_v^p &= \text{SelfAttentionPooling}(\mathbf{h}_v'), \\
\mathbf{h}_a^p &= \text{SelfAttentionPooling}(\mathbf{h}_a'), \\
\mathbf{h}_t^p &= \text{SelfAttentionPooling}(\mathbf{h}_t').
\end{aligned}
\end{equation}

Here, $\mathbf{h}_v^p$, $\mathbf{h}_a^p$, and $\mathbf{h}_t^p$ are the pooled representations for the visual, auditory, and textual modalities. By applying multi-head attention, the self-attention pooling mechanism ensures that the model prioritizes contextually important elements in each sequence.

\subsubsection{Cross-Attention}
To capture cross-modal dependencies, we employ multi-head cross-attention mechanisms, where each modality aligns with and incorporates information from the other two. For instance, the visual representation $\mathbf{h}_v^p$ attends to the audio $\mathbf{h}_a^p$ and text $\mathbf{h}_t^p$ modalities as follows:
\begin{equation}
\mathbf{h}_v^{ca} = \text{CrossAttention}(\mathbf{h}_v^p, \{\mathbf{h}_a^p, \mathbf{h}_t^p\})
\end{equation}

Similarly, the audio and text modalities are cross-attended using visual and textual embeddings, or visual and audio embeddings, respectively:

\begin{equation}
\begin{aligned}
\mathbf{h}_a^{ca} &= \text{CrossAttention}(\mathbf{h}_a^p, \{\mathbf{h}_v^p, \mathbf{h}_t^p\}), \\
\mathbf{h}_t^{ca} &= \text{CrossAttention}(\mathbf{h}_t^p, \{\mathbf{h}_v^p, \mathbf{h}_a^p\}).
\end{aligned}
\end{equation}

Each cross-attended output $\mathbf{h}_v^{ca}$, $\mathbf{h}_a^{ca}$, and $\mathbf{h}_t^{ca}$ combines modality-specific features with contextual information from the other modalities. This step allows the model to simulate human-like sensory integration by combining complementary information across modalities.
\subsubsection{Fusion Network}
The cross-attended embeddings are concatenated into a unified representation:

\begin{equation}
\mathbf{f} = [\mathbf{h}_v^{ca}, \mathbf{h}_a^{ca}, \mathbf{h}_t^{ca}]
\end{equation}

This fused embedding $\mathbf{f}$ is passed through a fully connected fusion network with ReLU activations and dropout layers. The network reduces the dimensionality to produce a single memorability score for each advertisement:

\begin{equation}
\hat{y} = \text{Sigmoid}(\text{Linear}(\text{ReLU}(\text{Dropout}(\mathbf{f}))))
\end{equation}

The final output $\hat{y}$ is a scalar value between 0 and 1, representing the predicted memorability score.

\subsection{Advertisement Regeneration
}
To examine the potential application MindMem, we generate more memorable advertisements for the commercial market. To achieve this, we gathered two types of information from the LAMBDA dataset (2,000 advertisements): (1) general video description and (2) scene-specific description. General video description represents details about the video itself, such as the brand name, advertisement orientation, advertisement pace, sentiment and audio description. Scene-specific features encompass detailed elements like scene descriptions, the emotion or mood of each scene, associated tags, dominant color theme, photography style, on-screen text, and the overall tone of each scene. Gemini Pro 1.5 was used to get those descriptions (prompt is shown in Appendix 1).  We fine-tuned the LLaMA 3.1 (8B) \cite{dubey2024llama} model using titles and key messages from the advertisements as input. The output of the model was to generate detailed descriptions of the advertisements, closely aligned with the output structure of Gemini Pro 1.5. We refer to the model developed through this process as MindMem-ReAd, designed to enhance the creation of highly memorable advertisements. MindMem-ReAd generated textual descriptions of individual scenes, incorporating both general video and scene-specific features.

\section{Result}
\subsection{Performance of MindMem to Predict Advertisement Memorability
}
We trained MindMem models using the LAMBDA training set, constructing models with varying modalities (single-modal, dual-modal, and multimodal) and incorporating different advanced attention mechanisms in the multimodal models, such as multi-head self-attention for capturing intra-modal dependencies and multi-head cross-attention layers for integrating and aligning information across modalities. In total, we developed 11 models and compared their performance to predict memorability on the LAMBDA test set. Then, we further compared the performance of MindMem with those of other cutting-edge methods such as Henry \cite{harinilong}, ViT-Mem\cite{hagen2023image}, GPT 3.5 and GPT 4O \cite{achiam2023gpt}. 
As shown in Table~\ref{tab:performance_comparison}, MindMem outperformed both single- and dual-modal models. MindMem achieved a Spearman’s correlation coefficient ($\rho = 0.631$) with statistical significance ($p$-value = $1.26 \times 10^{-13}$), improving $\rho$ by an average of 21\% compared to single-modal models and by 5\% compared to dual-modal models. It also showed the smallest mean squared error (Mean Squared Error, MSE = 0.048), indicating strong correlation between predictive and actual memorability scores. Among the single-modal approaches, the text-based model performed best with $\rho = 0.589$ (MSE = 0.062). Single audio was not enough by itself to produce good results. For dual-modal models, the combination of textual and video information yielded the highest performance, with $\rho = 0.615$ (MSE = 0.053). Meanwhile, three single-modal models showed relatively lower performance underscoring the limitation of relying on a single modality for memorability prediction. These results support the importance of a multimodal approach in capturing the intricate dynamics of human memory, particularly in memorability prediction.

\begin{table}[t]
    \centering
    \caption{Performance comparison among single-, double-, and multimodal models.}
    \resizebox{\linewidth}{!}{%
    \begin{tabular}{lcc}
        \hline
        {Modality} & {Spearman’s $\rho$ } & {MSE} \\
        \hline
        Video only          & 0.564                         & 0.057 \\
        Text only           & 0.589                    & 0.062 \\
        Audio only          & 0.336                        & 0.068 \\
        Text + Audio        & 0.605                       & 0.057 \\
        Text + Video        & 0.615                       & 0.053 \\
        Audio + Video       & 0.590                       & 0.054 \\
        {MindMem (Audio + Video + Text)} 
                            & \textbf{0.631} & \textbf{0.048} \\
        \hline
    \end{tabular}}
    \label{tab:performance_comparison}
    \vspace{0.5em}

\end{table}

\begin{figure}[t]
    \centering
    \includegraphics[width=0.87\linewidth]{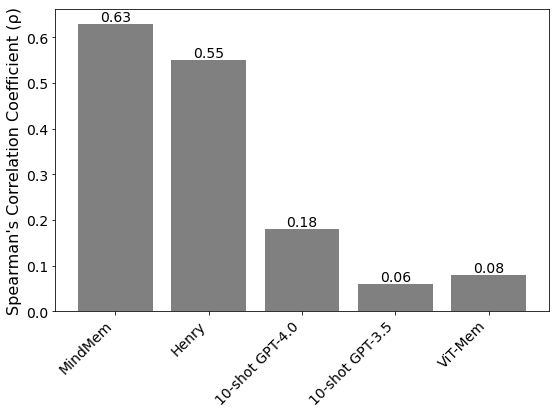} 
    \caption{Performance comparison of MindMem with four state-of-the-art methods: Henry, 10-shot GPT3.5, 10-shot GPT4.0-o, and Vit-Mem. MindMem consistently outperformed the others, achieving the highest average accuracy and Spearman’s correlation coefficient ($\rho$).}
    \label{fig:performance_comparison}
\end{figure}

We also compared MindMem’s performance with four other state-of-the-art methods: Henry, 10-shot GPT3.5, 10-shot GPT4.0-o, and Vit-Mem (Figure~\ref{fig:performance_comparison}). MindMem consistently outperformed the other methods, achieving an average accuracy that was 6.5 times higher. Notably, it outperformed Henry (the best-performing method among the others) by 13\% in terms of Spearman’s correlation coefficient ($\rho$).

\subsection{Ablation Study 
}
We conducted ablation tests to evaluate the impact of different architectural choices, such as various pooling and attention methods. As summarized in Table~\ref{table:mindmem_ablation}, simpler methods, like average pooling ($\rho = 0.424$, MSE = 0.079) and max pooling ($\rho = 0.462$, MSE = 0.071), showed the weakest performance. In contrast, advanced attention methods such as self-attention ($\rho = 0.614$, MSE = 0.052), cross-attention with average pooling ($\rho = 0.526$, MSE = 0.066), and a combination of self- and cross-attention layers ($\rho = 0.631$, MSE = 0.048) significantly improved predictions.

The results in Table~\ref{table:mindmem_ablation} indicate that while basic pooling has limitations in capturing contextual information, neuro-inspired methods (i.e., advanced attention methods) are more effective at extracting relevant multimodal features, leading to enhanced model performance.
\begin{table}[t]
    \centering
    \renewcommand{\arraystretch}{1.3} 
    \caption{Ablation tests on the MindMem model variations. The symbols $\checkmark$ and $\times$ indicate the inclusion or exclusion of features, respectively.}
    \label{table:mindmem_ablation}
    \resizebox{\linewidth}{!}{%
    \begin{tabular}{cccccc}
        \hline
        {Self-Attention} & {Cross-Attention} & {Average Pooling} & {Max Pooling} & {$\rho$} & {MSE} \\
        \hline
        $\times$ & $\times$ & $\checkmark$ & $\times$ & 0.424 & 0.079 \\
        $\checkmark$ & $\times$ & $\times$ & $\times$ & 0.614 & 0.052 \\
        $\times$ & $\checkmark$ & $\checkmark$ & $\times$ & 0.526 & 0.066 \\
        $\times$ & $\times$ & $\times$ & $\checkmark$ & 0.462 & 0.071 \\
        $\checkmark$ & $\checkmark$ & $\times$ & $\times$ & \textbf{0.631} & \textbf{0.048} \\
        \hline
    \end{tabular}}
\end{table}

\subsection{Content Factors Influencing Video Memorability}

Next, we investigated the relationship between content factors and memorability on the LAMBDA samples in the test set. We found a positive correlation between predicted memorability and video pace (overall video speed, rhythm, tone, or flow at which the content of a video unfolds). Videos with a higher pace tend to be remembered for a longer duration by the audience. High-paced videos had an average memorability score of $0.672 \pm 0.221$, whereas low-paced videos scored $0.499 \pm 0.229$, reflecting about 30\% lower memorability for slower-paced videos with a statistical significance ($p$-value = $8.32 \times 10^{-4}$, one-way ANOVA test; Figure 3a). 

The number of scenes in an advertisement also exhibited a positive relationship with memorability. Advertisements with a greater number of scenes were remembered for longer durations by audiences ($p$-value = $5.12 \times 10^{-5}$, one-way ANOVA test; Figure 3b). Interestingly, advertisements that evoked more emotions were significantly more memorable ($\rho = 0.366$, $p$-value = $1.29 \times 10^{-7}$; Figure 3c).

In contrast, factors such as the orientation of the advertisement (portrait vs. landscape; Figure 3d), the advertisement's duration (Figure 3e), and the number of color themes(Figure 3f) showed an insignificant relationship with memorability scores (p-value $>$ 0.05).

\begin{figure}[t]
    \centering
    \includegraphics[width=0.5\textwidth]{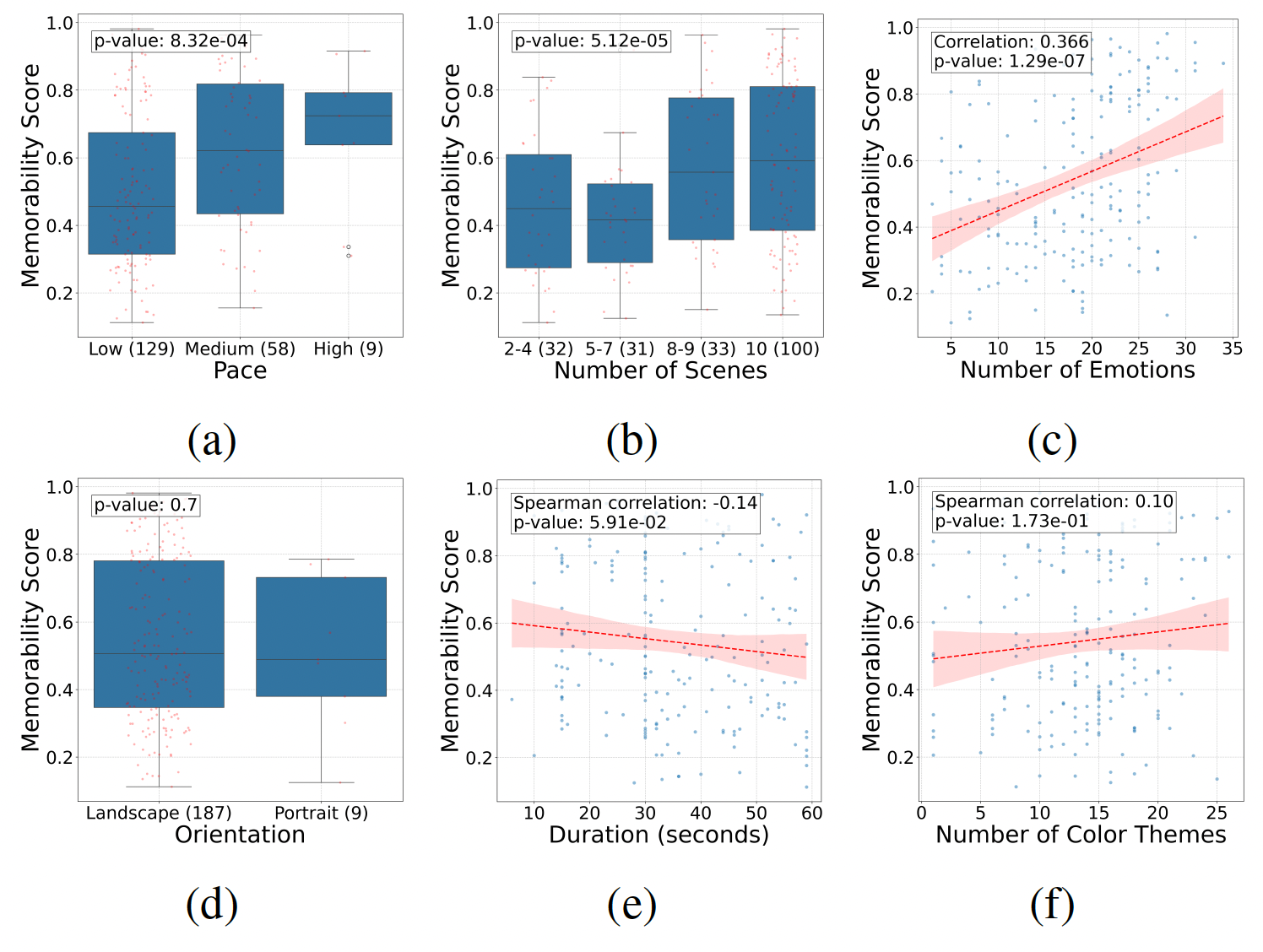} 
    \caption{Relationship between content factors and memorability scores on the LAMBDA samples in a test set. (a) video pace, (b) number of scenes, (c) number of emotions in a video, (d) video orientation, (e) video duration and (f) number of color themes are compared with predicted memorability scores. Statistical significance is measured using one-way ANOVA test (a and b), and T-test (d). Spearman’s correlation coefficient is displayed for scatter plots (c, e, and f).}
    \label{fig:multi_panel)}
    \label{fig:Figure 3}
\end{figure}

\subsection{MindMem Architecture Validation}

To further evaluate the reliability of the MindMem architecture, we conducted experiments using the Memento10K dataset. Unlike the LAMBDA dataset, Memento10K features distinct characteristics, consisting of relatively short (3-second) natural videos. A total of 7,000 videos were used to train the MindMem model, and it was evaluated on a validation set of 1,500 videos. We observed that the MindMem architecture is stable and consistently delivers reliable prediction performance across various datasets. Specifically, in models based on Memento10K, MindMem achieved a Spearman's correlation coefficient ($\rho$) of 0.731 (MSE = 0.0055) when all three types of multimodal information were fed into the model (Table~\ref{tab:mindmem_performance}). The dual-modal model combining text and video information demonstrated similar performance ($\rho = 0.728$) to MindMem's. We suspect that the 3-second audio clips provide insufficient information for accurate memorability predictions. Indeed, single-modal approaches showed the lowest performance, with the audio-only model achieving a $\rho$ of 0.291, making it the poorest performance among the three.

We also compared the performance of our model with other models that tested the Memento10K dataset. MindMem demonstrated superior results, outperforming the other methods (Table~\ref{tab:comparison_sota}) and achieving an average of 1.3 times higher accuracy in predicting memorability.

\begin{table}[h]
\centering
\caption{The performance of MindMem on the Memento10K dataset.}
\label{tab:mindmem_performance}
\begin{tabular}{lcc}
\hline
{Modality} & {Spearman's $\rho$} & {MSE} \\ \hline
Video only & 0.709 & 0.006 \\
Text only & 0.648 & 0.007 \\
Audio only & 0.291 & 0.012 \\
Text + Audio & 0.682 & 0.006 \\
Text + Video & 0.728 & 0.006 \\
Audio + Video & 0.697 & 0.006\\
MindMem (Audio + Video + Text) & \textbf{0.731} & \textbf{0.006 } \\ \hline
\end{tabular}
\end{table}

\begin{table}[t]  
\centering
\footnotesize     
\resizebox{\columnwidth}{!}{
\begin{tabular}{lc}
\hline
\textbf{Approach} & \textbf{Spearman's $\rho$} \\
\hline
MemNet baseline \cite{khosla2015understanding}        & 0.485 \\
M3-S \cite{dumont2023modular}                         & 0.670 \\
SemanticMemNet \cite{newman2020multimodal}            & 0.659 \\
Cohendet et al.\ (ResNet3D) \cite{cohendet2019videomem} & 0.574 \\
Cohendet et al.\ (Semantic) \cite{cohendet2019videomem}& 0.552 \\
\textbf{MindMem (All 3)}                             & \textbf{0.731} \\
\hline
\end{tabular}
}
\caption{Comparison to state-of-the-art on Memento10K.}
\label{tab:comparison_sota}
\end{table}

\subsection{Generating Memorable Advertisements
}
\subsubsection{Quantitative Evaluation of Advertisement Regeneration}
We investigated the practical application of MindMem-driven memorability prediction by targeting the creation of more memorable advertisements for the commercial market. To achieve this, we developed MindMem-ReAd, a system built by fine-tuning the LLaMA 3.1 (8B) model to simulate advertisement content and predict memorability scores. We applied MindMem-ReAd to a set of 50 commercial advertisements. These videos were randomly selected from YouTube and represent 10 diverse industries, including food and beverage, technology and gadgets, beauty and personal care, health and wellness, fashion and apparel, automotive, entertainment and media, travel and hospitality, home and living, and finance and insurance.

To evaluate the effectiveness of MindMem-ReAd, we assessed both the original and the regenerated advertisements using our text-only trained model as an objective measure of memorability. By using the text-only model as a judge, we were able to predict memorability scores for the advertisements based solely on their textual content, allowing us to directly compare the impact of MindMem-ReAd on enhancing advertisement memorability.

MindMem-ReAd improved overall 19.14\% of memorability compared to original advertisements (Table~\ref{tab:memorability_improvement}). Of 50 tested advertisements, 16 had an original memorability score of $\leq 0.5$, representing low-memorable advertisements that demonstrated an average improvement of 74.12\%. Additionally, advertisements with medium memorability scores ($0.5 < \text{original memorability} < 0.7$) and high memorability scores ($\text{original memorability} \geq 0.7$) showed improvements of 14.82\% and 2.13\%, respectively.
\begin{table}[h]
\centering
\caption{Performance of MindMem-ReAd on advertisements across different memorability categories. 16 low, 18 medium, and 16 high memorable videos are used for the analysis. Overall indicates combined performance across all categories (50 videos). Improvement is expressed as the percentage increase in memorability scores achieved by MindMem-ReAd compared to the original scores.}

\label{tab:memorability_improvement}
\setlength{\tabcolsep}{1.3pt} 
\renewcommand{\arraystretch}{1.2} 
\begin{tabular}{lccc}
\hline
{Category} & {Original} & {MindMem-ReAd} & { Improvement} \\ \hline
Low & $0.340 \pm 0.099$ & $0.592 \pm 0.137$ & 74.12\% \\
Medium & $0.614 \pm 0.046$ & $0.705 \pm 0.082$
 & 14.82\% \\
High & $0.846 \pm 0.063$ & $0.864 \pm 0.077$ & 2.13\% \\ 
Overall & $0.606 \pm 0.215$ & $0.722 \pm 0.148$ & 19.14\% \\ \hline
\end{tabular}
\end{table}

\subsubsection{Case Studies}

We provide a detailed analysis of two re-generated advertisements as case studies to demonstrate our approach. The evaluation focuses on four key metrics that assess whether the memorable advertisement holds greater marketing appeal or impact on general audiences: (1) memorability score predicted by the single-modal text model, (2) clarity, (3) visual impact, and (4) customer retention, assessed using GPT-o1-preview and Perplexity. 

\begin{figure}[t]
    \centering
    \includegraphics[width=9cm]{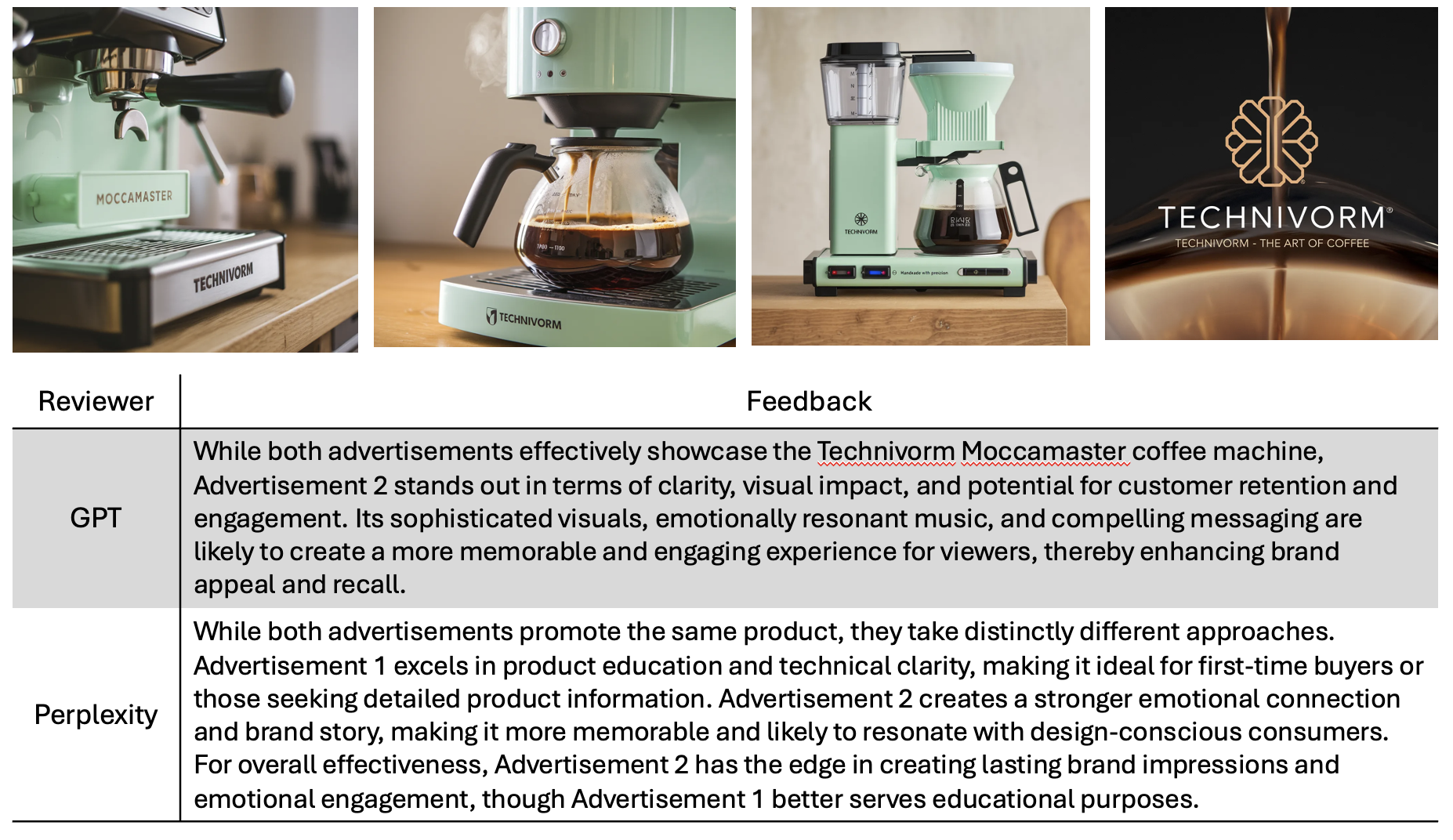} 
    \caption{Advertisement generated by MindMem-ReAd (Advertisement \#1). Images are created using Ideogram (https://ideogram.ai)}
    \label{fig:usecase_1}
\end{figure}

\textbf{i. Advertisement \#1: Technivorm Moccamaster Coffee Machine} \\
The original version of Advertisement \#1 achieved a memorability score of 0.19. In contrast, the MindMem-ReAd advertisement attained a significantly higher memorability score of 0.62, reflecting an improvement of more than 3-fold. According to the evaluation from GPT-o1-preview and Perplexity, the MindMem-ReAd advertisement excels in clarity, visual impact, and its potential to enhance customer retention and engagement (Figure \ref{fig:usecase_1} and Appendix~2).

\begin{figure}[t]
    \centering
    \includegraphics[width=9cm]{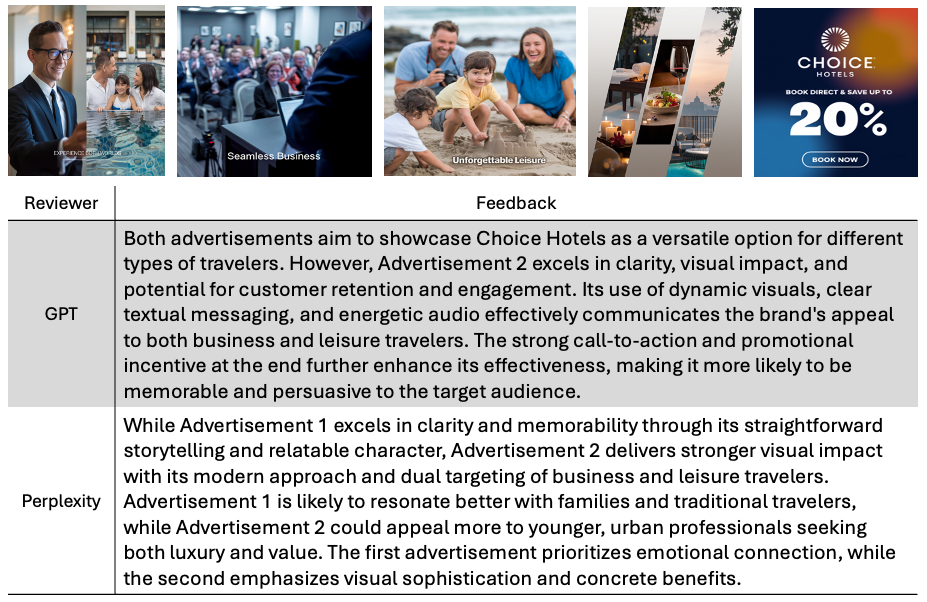} 
    \caption{Advertisement generated by MindMem-ReAd (Advertisement \#2). Images are created using Ideogram (https://ideogram.ai)}
    \label{fig:usecase_2}
\end{figure}

\textbf{ii. Advertisement \#2: Choice Hotels} \\
The original advertisement achieved a memorability score of 0.23, while the MindMem-ReAd version excelled with a score of 0.46, marking an improvement of 2-fold. Feedback from GPT-o1-preview and Perplexity commonly highlighted the MindMem-ReAd output for its effective use of dynamic visuals, clear textual messaging, and energetic audio, which successfully conveyed the brand's appeal to both business and leisure travelers, resulting in greater impact and broader retention (Figure \ref{fig:usecase_2} and Appendix~3). These results underscore the potential of MindMem not only to predict but also to create highly memorable advertising content.

\section{Discussion
}

\subsection{Our Contributions}
Deep learning algorithms and large language models have the potential to transform commercial advertisement generation by enhancing strategic interactions for advertisers within multi-agent settings. Focusing on the advertiser's role in these complex interactions, our research aims to optimize advertising strategies—specifically in predicting and enhancing advertisement memorability. We introduce MindMem, a multimodal framework that utilizes advanced attention mechanisms on textual, visual, and auditory data to achieve high accuracy in predicting memorability, which is a key aspect of strategic communication. To demonstrate real-world applicability, we developed MindMem-ReAd, an LLM-driven system that optimizes advertisement content to enhance memorability and boost consumer engagement. This work bridges memorability prediction with practical multi-agent advertising strategies, highlighting the potential of generative AI to drive targeted and impactful marketing campaigns.

\subsection{The Impact of Neuro-inspired Approaches On Predictive Performance
}
The incorporation of neuro-inspired mechanisms, particularly advanced attention models, has been instrumental in enhancing the predictive performance of MindMem. Drawing inspiration from human cognitive processes, these mechanisms enable the model to simulate how the brain selectively focuses on and integrates multimodal information, thereby improving its ability to predict advertisement memorability.

In the architecture of MindMem, we implemented multi-head self-attention pooling and cross-attention layers to capture both intra-modal and inter-modal dependencies. The self-attention pooling mechanism allows the model to weigh the importance of different elements within each modality's sequence, akin to how human attention selectively prioritizes certain stimuli over others. This is crucial for handling variable-length sequences and emphasizing contextually relevant features within the visual, auditory, and textual data.

The cross-attention layers further enhance this capability by enabling the model to align and integrate information across different modalities. This mirrors the human brain's ability to synthesize sensory information from various sources to form a coherent perception of an event or scene. By allowing each modality to attend to the others, the model captures complex interactions and dependencies that are essential to predict memorability, which is inherently a multimodal cognitive function.

\subsection{Relationship between Video Dynamics and Memory Formation}
We observed a positive correlation between memorability and dynamic content factors such as video pace and diversities of scene and emotion. The positive correlation between video pace and memorability aligns with recent research showing that faster-paced content can lead to better engagement and information retention \cite{murphy2022learning}. Similarly, incorporating a greater number of scenes contributes to a faster-paced video, fostering sustained interest and offering more cognitive hooks to aid memory retention. We suspect that the relationship between emotional diversity and memorability represents a complex interaction of cognitive and emotional processes. It has been suggested that higher emotional diversity, characterized by the richness and balance of emotional experience is associated with improved cognitive functioning \cite{urban2022emodiversity}. This enhanced cognitive state would facilitate more effective memory encoding and retrieval processes. Meanwhile, traditional design elements, such as advertisement orientation, color themes, and advertisement duration, showed insignificant relationship with memorability, suggesting that content richness and emotional resonance would outweigh static structural features. These findings highlight the need for advertisers to prioritize dynamic and emotionally engaging elements over conventional design considerations to create more impactful and memorable advertisements.

\subsection{Limitation}
While our study demonstrates the effectiveness of neuro-inspired techniques in improving memorability prediction, several limitations remain. First, the size and variety of data, though substantial, may still limit the generalizability of our findings across different types of advertisements and industries. Additionally, our models rely on specific multimodal inputs, which might not capture other relevant factors like cultural context or individual-specific biases that could influence memorability. Future research require investigation of more diverse datasets and consideration of broader contextual factors, such as language variations, cultural diversity, or individual preferences. Additionally, integrating more advanced neuro-inspired mechanisms could further refine the model's ability to mimic human cognitive processes, potentially improving predictive accuracy and explainability.

\section{Conclusion
}
In this study, we presented MindMem, a multimodal model designed to predict advertisement memorability. MindMem mimicked human cognitive processes, and significantly enhanced the model’s ability to integrate visual, auditory, and textual inputs, leading to more accurate predictions compared to currently available other models. In addition, we developed MindMem-ReAd, a scalable method to generate memorable advertisements that achieved significantly higher memorability scores than their original versions. These findings underscore the potential of combining generative AI with cognitive modeling to optimize advertising strategies and enhance consumer engagement. Future work will focus on extending this framework by integrating more diverse datasets, applying it to practical advertisement content generation in multi-agent strategic settings, and exploring additional cognitive mechanisms to further improve model performance and broaden its applicability.

\bibliographystyle{aaai25}
\bibliography{aaai25}

\begin{thebibliography}{21}
\providecommand{\natexlab}[1]{#1}

\bibitem[{Achiam et~al.(2023)Achiam, Adler, Agarwal, Ahmad, Akkaya, Aleman, Almeida, Altenschmidt, Altman, Anadkat et~al.}]{achiam2023gpt}
Achiam, J.; Adler, S.; Agarwal, S.; Ahmad, L.; Akkaya, I.; Aleman, F.~L.; Almeida, D.; Altenschmidt, J.; Altman, S.; Anadkat, S.; et~al. 2023.
\newblock Gpt-4 technical report.
\newblock \emph{arXiv preprint arXiv:2303.08774}.

\bibitem[{Chu et~al.(2024)Chu, Xu, Yang, Wei, Wei, Guo, Leng, Lv, He, Lin, Zhou, and Zhou}]{Qwen2-Audio}
Chu, Y.; Xu, J.; Yang, Q.; Wei, H.; Wei, X.; Guo, Z.; Leng, Y.; Lv, Y.; He, J.; Lin, J.; Zhou, C.; and Zhou, J. 2024.
\newblock Qwen2-Audio Technical Report.
\newblock \emph{arXiv preprint arXiv:2407.10759}.

\bibitem[{Cohendet et~al.(2019)Cohendet, Demarty, Duong, and Engilberge}]{cohendet2019videomem}
Cohendet, R.; Demarty, C.-H.; Duong, N.~Q.; and Engilberge, M. 2019.
\newblock VideoMem: Constructing, analyzing, predicting short-term and long-term video memorability.
\newblock In \emph{Proceedings of the IEEE/CVF International Conference on Computer Vision}, 2531--2540.

\bibitem[{Dubey et~al.(2024)Dubey, Jauhri, Pandey, Kadian, Al-Dahle, Letman, Mathur, Schelten, Yang, Fan et~al.}]{dubey2024llama}
Dubey, A.; Jauhri, A.; Pandey, A.; Kadian, A.; Al-Dahle, A.; Letman, A.; Mathur, A.; Schelten, A.; Yang, A.; Fan, A.; et~al. 2024.
\newblock The llama 3 herd of models.
\newblock \emph{arXiv preprint arXiv:2407.21783}.

\bibitem[{Dubey et~al.(2015)Dubey, Peterson, Khosla, Yang, and Ghanem}]{7410487}
Dubey, R.; Peterson, J.; Khosla, A.; Yang, M.-H.; and Ghanem, B. 2015.
\newblock What Makes an Object Memorable?
\newblock In \emph{2015 IEEE International Conference on Computer Vision (ICCV)}, 1089--1097.

\bibitem[{Dudzik et~al.(2020)Dudzik, Hung, Neerincx, and Broekens}]{dudzik2020investigating}
Dudzik, B.; Hung, H.; Neerincx, M.; and Broekens, J. 2020.
\newblock Investigating the influence of personal memories on video-induced emotions.
\newblock In \emph{Proceedings of the 28th ACM conference on user modeling, adaptation and personalization}, 53--61.

\bibitem[{Dumont, Hevia, and Fosco(2023)}]{dumont2023modular}
Dumont, T.; Hevia, J.~S.; and Fosco, C.~L. 2023.
\newblock Modular memorability: Tiered representations for video memorability prediction.
\newblock In \emph{Proceedings of the IEEE/CVF Conference on Computer Vision and Pattern Recognition}, 10751--10760.

\bibitem[{Hagen and Espeseth(2023)}]{hagen2023image}
Hagen, T.; and Espeseth, T. 2023.
\newblock Image memorability prediction with vision transformers.
\newblock \emph{arXiv preprint arXiv:2301.08647}.

\bibitem[{HariniSI et~al.(2024)HariniSI, Singh, Singla, Bhattacharyya, Baths, Chen, Shah, and Krishnamurthy}]{harinilong}
HariniSI; Singh, S.; Singla, Y.~K.; Bhattacharyya, A.; Baths, V.; Chen, C.; Shah, R.~R.; and Krishnamurthy, B. 2024.
\newblock Long-Term Ad Memorability: Understanding and Generating Memorable Ads.
\newblock arXiv:2309.00378.

\bibitem[{Hu et~al.(2020)Hu, Huang, Li, Zhang, Zhang, and Liang}]{hu2020video}
Hu, W.; Huang, G.; Li, L.; Zhang, L.; Zhang, Z.; and Liang, Z. 2020.
\newblock Video-triggered EEG-emotion public databases and current methods: a survey.
\newblock \emph{Brain Science Advances}, 6(3): 255--287.

\bibitem[{Isola et~al.(2011)Isola, Parikh, Torralba, and Oliva}]{isola2011understanding}
Isola, P.; Parikh, D.; Torralba, A.; and Oliva, A. 2011.
\newblock Understanding the intrinsic memorability of images.
\newblock \emph{Advances in neural information processing systems}, 24.

\bibitem[{Khosla et~al.(2013)Khosla, Bainbridge, Torralba, and Oliva}]{khosla2013modifying}
Khosla, A.; Bainbridge, W.~A.; Torralba, A.; and Oliva, A. 2013.
\newblock Modifying the memorability of face photographs.
\newblock In \emph{Proceedings of the IEEE international conference on computer vision}, 3200--3207.

\bibitem[{Khosla et~al.(2015)Khosla, Raju, Torralba, and Oliva}]{khosla2015understanding}
Khosla, A.; Raju, A.~S.; Torralba, A.; and Oliva, A. 2015.
\newblock Understanding and predicting image memorability at a large scale.
\newblock In \emph{Proceedings of the IEEE international conference on computer vision}, 2390--2398.

\bibitem[{Li et~al.(2022)Li, Guo, Yue, Xue, and Sun}]{li2022adaptive}
Li, J.; Guo, X.; Yue, F.; Xue, F.; and Sun, J. 2022.
\newblock Adaptive Multi-Modal Ensemble Network for Video Memorability Prediction.
\newblock \emph{Applied Sciences}, 12(17): 8599.

\bibitem[{Murphy et~al.(2022)Murphy, Hoover, Agadzhanyan, Kuehn, and Castel}]{murphy2022learning}
Murphy, D.~H.; Hoover, K.~M.; Agadzhanyan, K.; Kuehn, J.~C.; and Castel, A.~D. 2022.
\newblock Learning in double time: The effect of lecture video speed on immediate and delayed comprehension.
\newblock \emph{Applied Cognitive Psychology}, 36(1): 69--82.

\bibitem[{Newman et~al.(2020)Newman, Fosco, Casser, Lee, McNamara, and Oliva}]{newman2020multimodal}
Newman, A.; Fosco, C.; Casser, V.; Lee, A.; McNamara, B.; and Oliva, A. 2020.
\newblock Multimodal memorability: Modeling effects of semantics and decay on video memorability.
\newblock In \emph{Computer Vision--ECCV 2020: 16th European Conference, Glasgow, UK, August 23--28, 2020, Proceedings, Part XVI 16}, 223--240. Springer.

\bibitem[{Team et~al.(2024)Team, Georgiev, Lei, Burnell, Bai, Gulati, Tanzer, Vincent, Pan, Wang et~al.}]{team2024gemini}
Team, G.; Georgiev, P.; Lei, V.~I.; Burnell, R.; Bai, L.; Gulati, A.; Tanzer, G.; Vincent, D.; Pan, Z.; Wang, S.; et~al. 2024.
\newblock Gemini 1.5: Unlocking multimodal understanding across millions of tokens of context.
\newblock \emph{arXiv preprint arXiv:2403.05530}.

\bibitem[{Urban-Wojcik et~al.(2022)Urban-Wojcik, Mumford, Almeida, Lachman, Ryff, Davidson, and Schaefer}]{urban2022emodiversity}
Urban-Wojcik, E.~J.; Mumford, J.~A.; Almeida, D.~M.; Lachman, M.~E.; Ryff, C.~D.; Davidson, R.~J.; and Schaefer, S.~M. 2022.
\newblock Emodiversity, health, and well-being in the Midlife in the United States (MIDUS) daily diary study.
\newblock \emph{Emotion}, 22(4): 603.

\bibitem[{Wang et~al.(2024)Wang, Jiang, Liu, Ma, Zhang, Pan, Liu, Gu, Xia, Li et~al.}]{wang2024comprehensive}
Wang, J.; Jiang, H.; Liu, Y.; Ma, C.; Zhang, X.; Pan, Y.; Liu, M.; Gu, P.; Xia, S.; Li, W.; et~al. 2024.
\newblock A comprehensive review of multimodal large language models: Performance and challenges across different tasks.
\newblock \emph{arXiv preprint arXiv:2408.01319}.

\bibitem[{Yang et~al.(2024)Yang, Yang, Hui, Zheng, Yu, Zhou, Li, Li, Liu, Huang et~al.}]{yang2024qwen2}
Yang, A.; Yang, B.; Hui, B.; Zheng, B.; Yu, B.; Zhou, C.; Li, C.; Li, C.; Liu, D.; Huang, F.; et~al. 2024.
\newblock Qwen2 technical report.
\newblock \emph{arXiv preprint arXiv:2407.10671}.

\bibitem[{Zhang et~al.(2024)Zhang, Zhang, Li, Zeng, Yang, Zhang, Wang, Tan, Li, and Liu}]{zhang2024long}
Zhang, P.; Zhang, K.; Li, B.; Zeng, G.; Yang, J.; Zhang, Y.; Wang, Z.; Tan, H.; Li, C.; and Liu, Z. 2024.
\newblock Long context transfer from language to vision.
\newblock \emph{arXiv preprint arXiv:2406.16852}.

\end{thebibliography}

\appendix
\onecolumn
\section{Appendix}
\subsection{Appendix 1: Gemini Pro 1.5 Prompt to Extract Textual Information from Advertisements.}

The following prompt was used with Gemini Pro 1.5 to extract detailed information from video advertisements:

\begin{lstlisting}
You are an advanced video analysis model tasked with extracting detailed information from a video advertisement.

Your goal is to identify and describe various elements of the video, including brand, core message, scenes (and their description), emotional appeal, mood, sound design, and other characteristics.

Please remember that if you do not know or if it is not in the video any of the below thing simply just say ""

It is important to get the result in json format.

Use the following format to organize your findings:

1- General Video Information:

Brand: Identify the brand associated with the video.

Orientation: Describe the video orientation (e.g., landscape, portrait).

Pace: Indicate the overall pace of the video (e.g., fast, slow).

Audio: Explain sound design in detail (e.g. sound effects or voiceovers or tone of voice) what is the sound in the video.

Sentiment: What is the sentiment of the video (e.g., positive, negative, neutral).

2- Scene Analysis (For each scene in the video, provide the following details):

Scene Number: Assign a number to each scene for reference.

Description: Provide a concise description of what happens in the scene.

Emotions or Mood: Identify the emotions conveyed by the scene (e.g., happy, tense).

Tags: List relevant keywords, objects, or tags associated with the scene.

Colors: Describe the dominant colors present in the scene.

Photography Style: Mention the photography style (e.g., close-up, wide shot).

Text Shown: Transcribe any text that appears on screen.

Tone: Describe the tone conveyed in the scene (e.g., formal, casual).

The output should look like:

{
  "General Video Information": {
    "Brand": "string",
    "Orientation": "string",
    "Pace": "string",
    "Audio": "string",
    "Sentiment": "string"
  },
  "Scene Analysis": [
    {
      "Scene Number": "integer",
      "Description": "string",
      "Emotions or Mood": "string",
      "Tags": ["string"],
      "Colors": ["string"],
      "Photography Style": "string",
      "Text Shown": "string",
      "Tone": "string"
    },
    {
      "Scene Number": "integer",
      "Description": "string",
      "Emotions or Mood": "string",
      "Tags": ["string"],
      "Colors": ["string"],
      "Photography Style": "string",
      "Text Shown": "string",
      "Tone": "string"
    }
    // Repeat the above structure for each scene in the video
  ]
}
\end{lstlisting}

\onecolumn
\subsection{Appendix 2: Memorability and Marketing Performance Indicator Comparison between Original and Re-generated Advertisement \#1.}

Original advertisement : https://www.youtube.com/watch?v=raH-O0AI8pQ

\begin{figure}[h!]
    \centering    \includegraphics[width=0.89\textwidth]{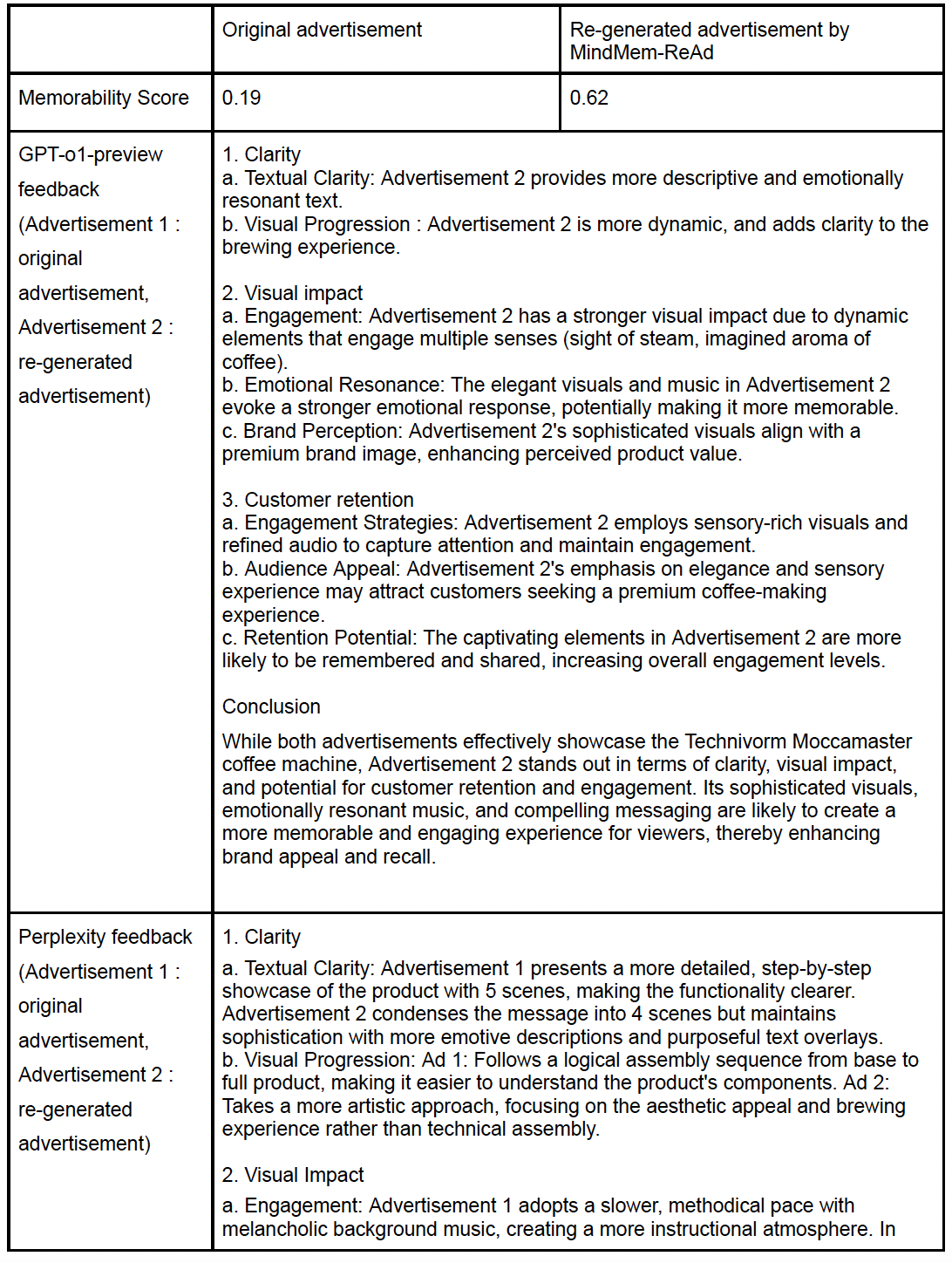}
\end{figure}

\begin{figure}[h!]
    \centering    \includegraphics[width=0.89\textwidth]{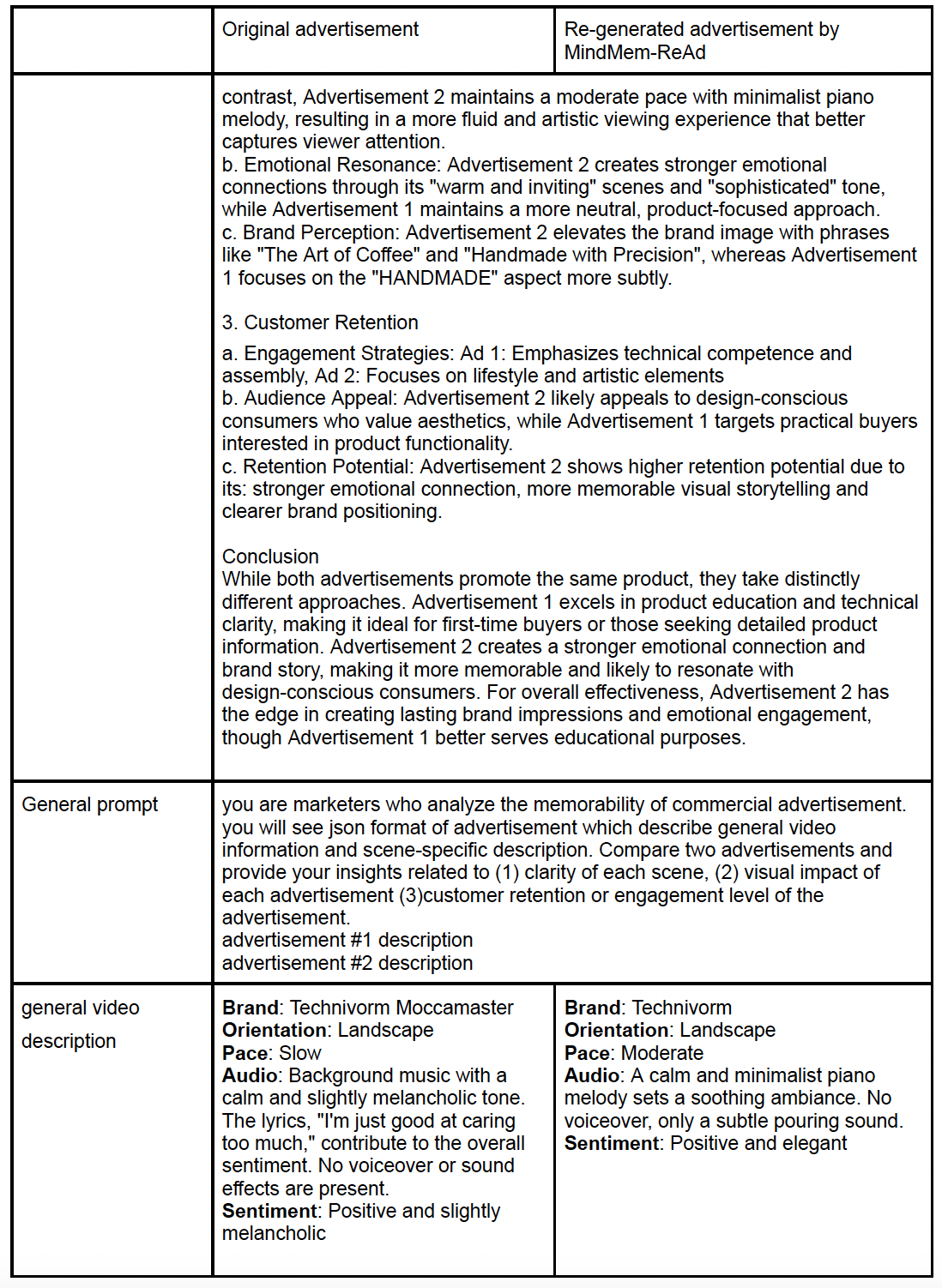}
\end{figure}

\begin{figure}[h!]
    \centering    \includegraphics[width=0.89\textwidth]{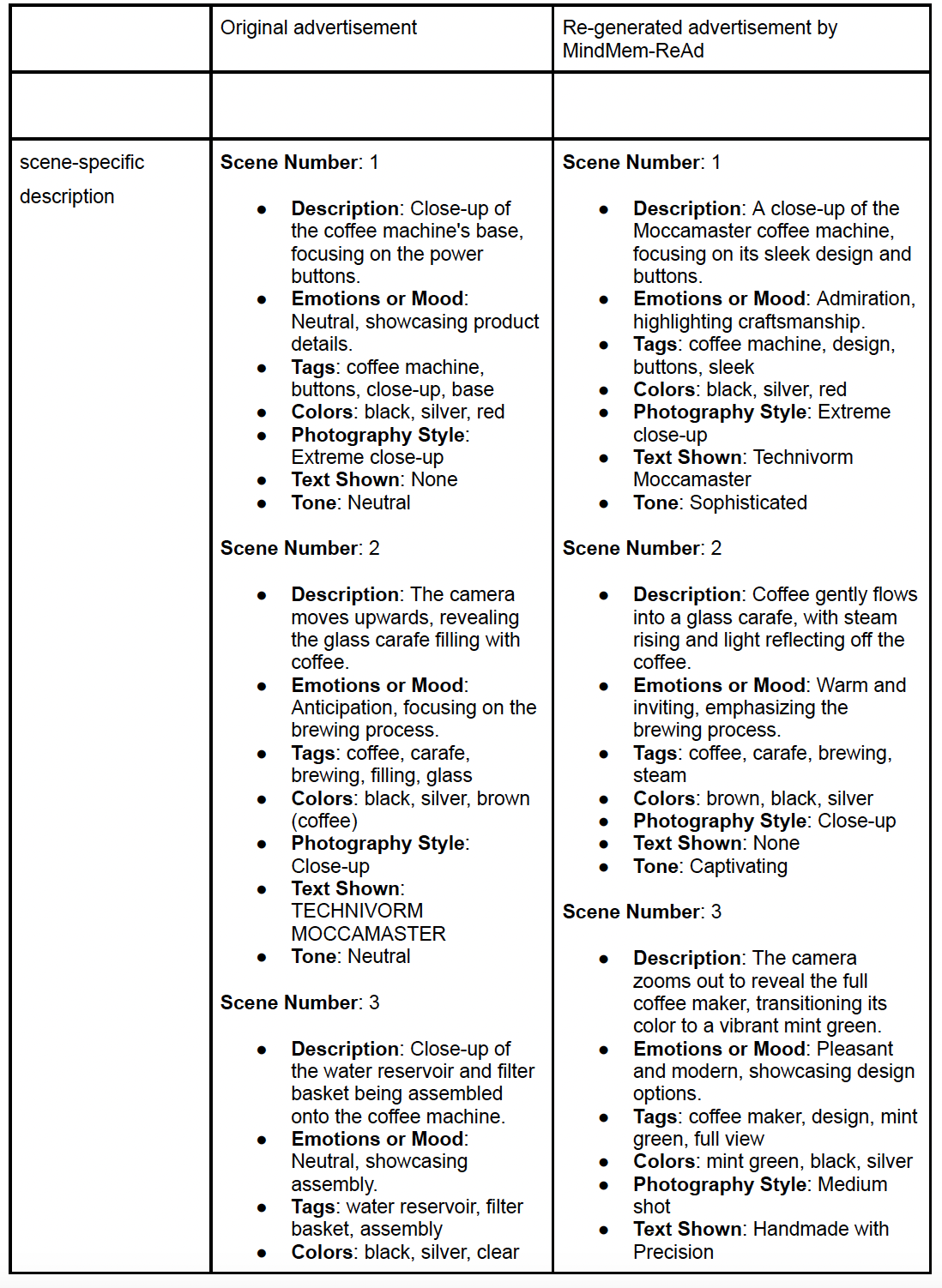}
\end{figure}

\begin{figure}[h!]
    \centering    \includegraphics[width=0.89\textwidth]{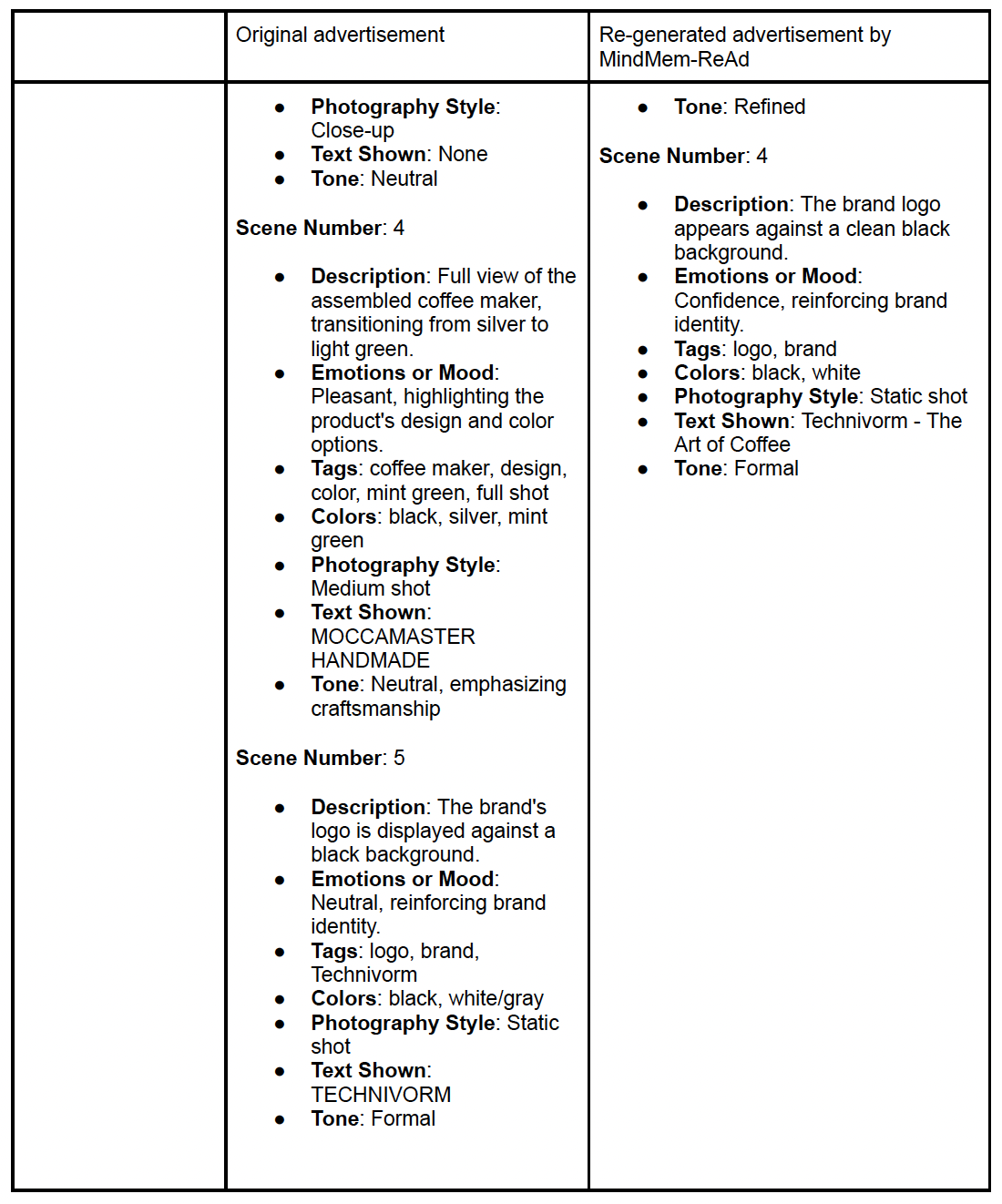}
\end{figure}

\subsection{Appendix 3: Memorability and Marketing Performance Indicator Comparison between Original and Re-generated Advertisement \#2.}

Original advertisement : https://www.youtube.com/watch?v=yj0xaRgRGaU

\begin{figure}[h!]
    \centering    \includegraphics[width=0.90\textwidth]{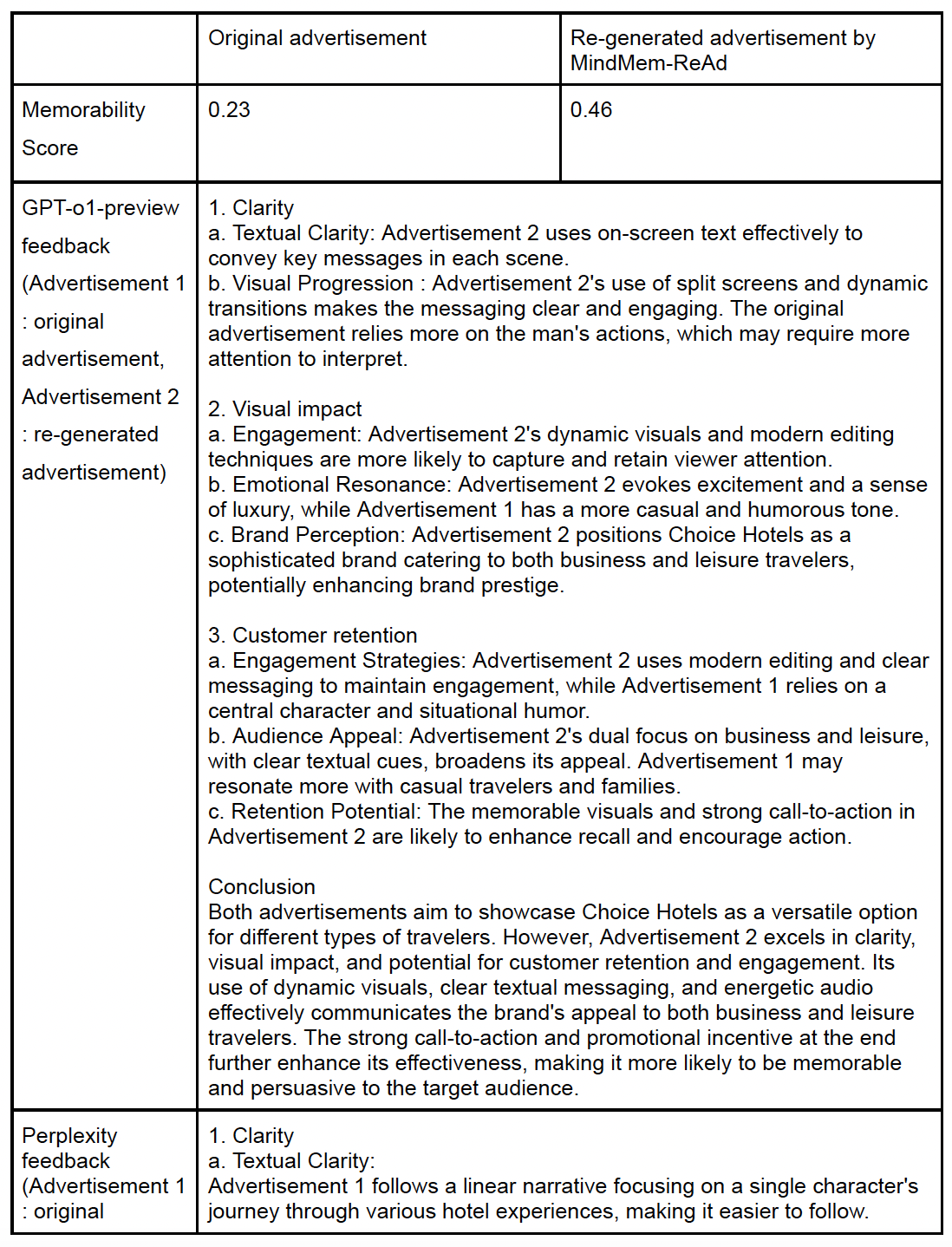}
\end{figure}

\begin{figure}[h!]
    \centering    \includegraphics[width=0.90\textwidth]{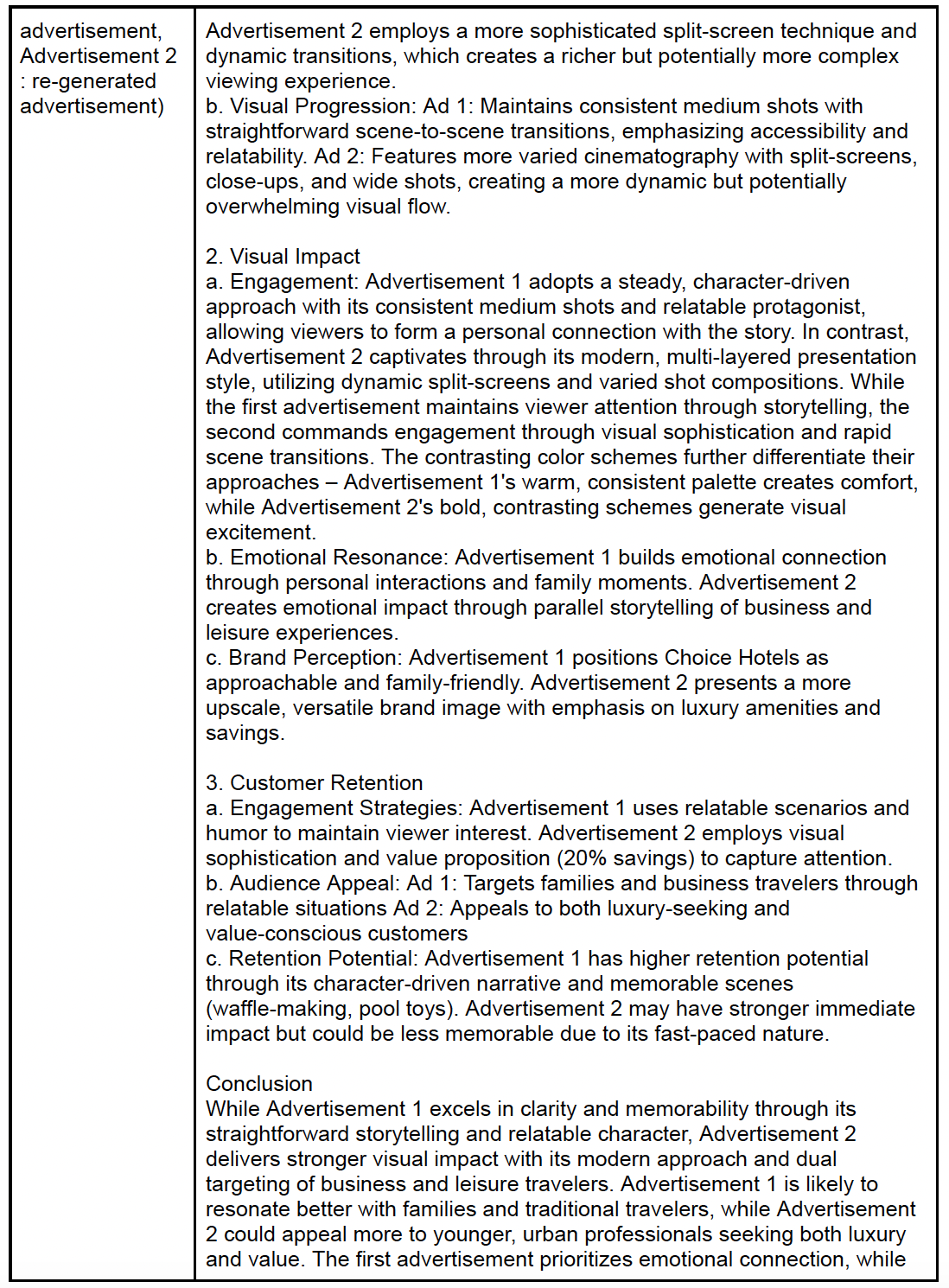}
\end{figure}

\begin{figure}[h!]
    \centering    \includegraphics[width=0.90\textwidth]{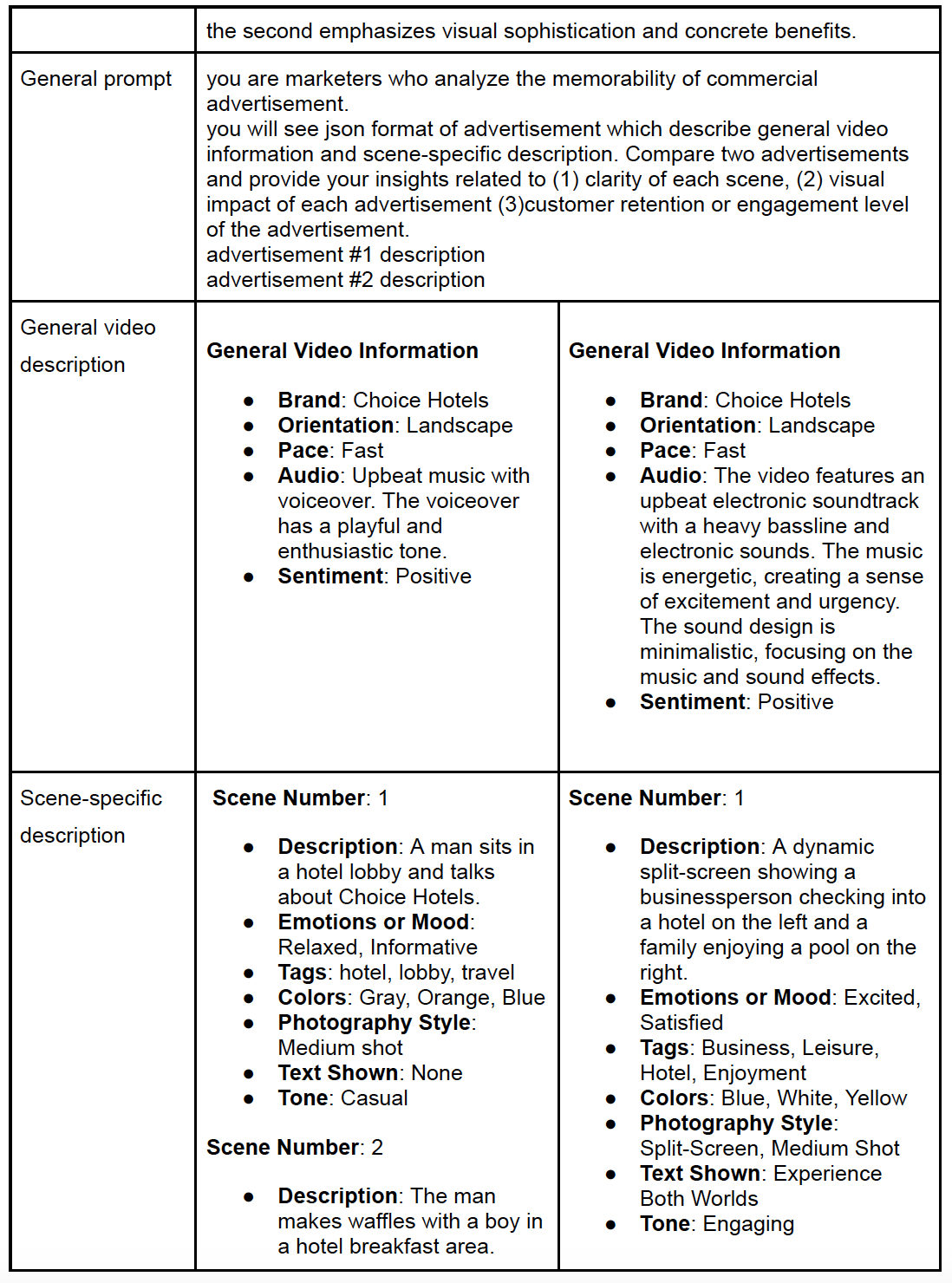}
\end{figure}

\begin{figure}[h!]
    \centering    \includegraphics[width=0.90\textwidth]{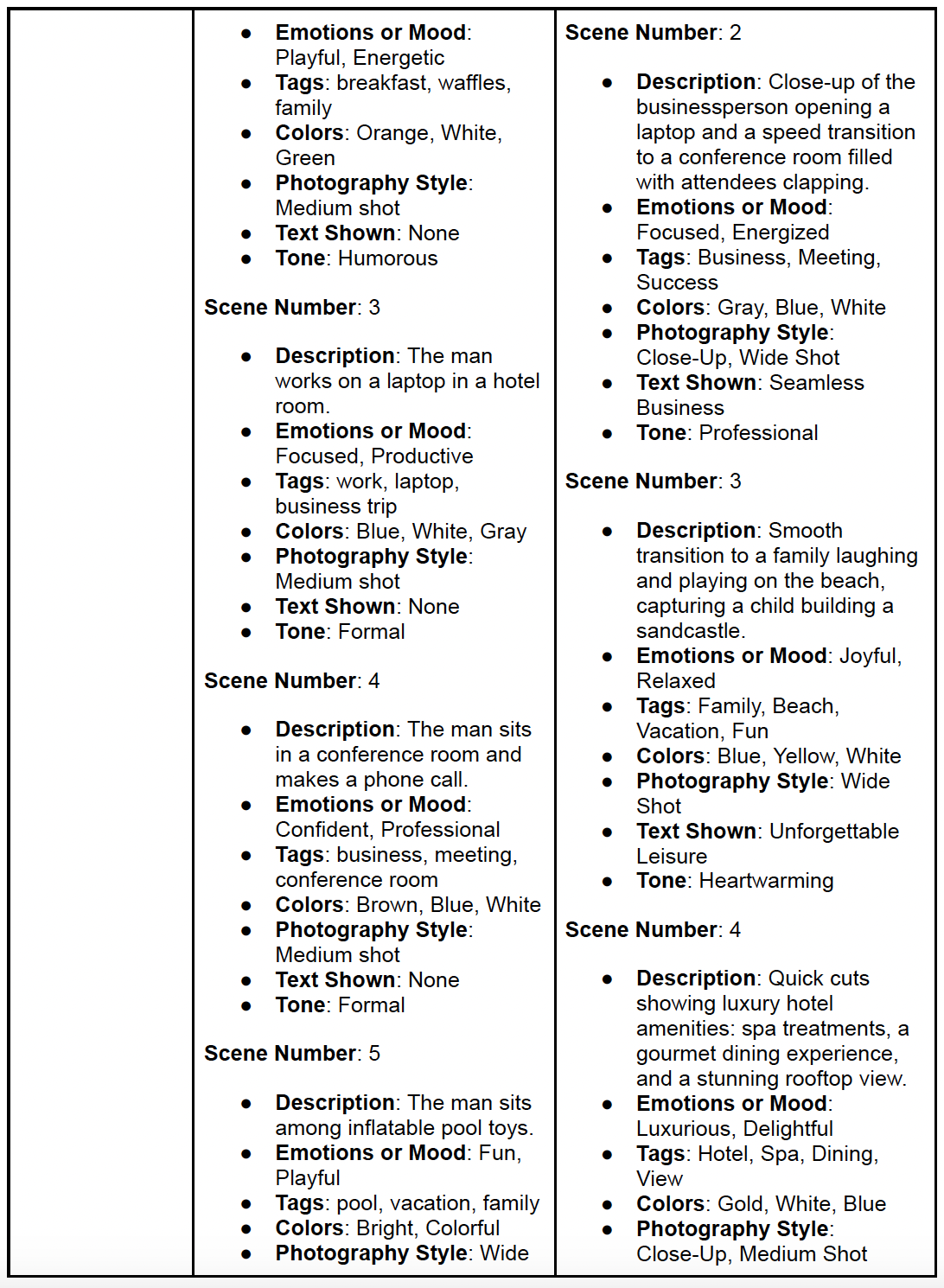}
\end{figure}

\begin{figure}[h!]
    \centering    \includegraphics[width=0.90\textwidth]{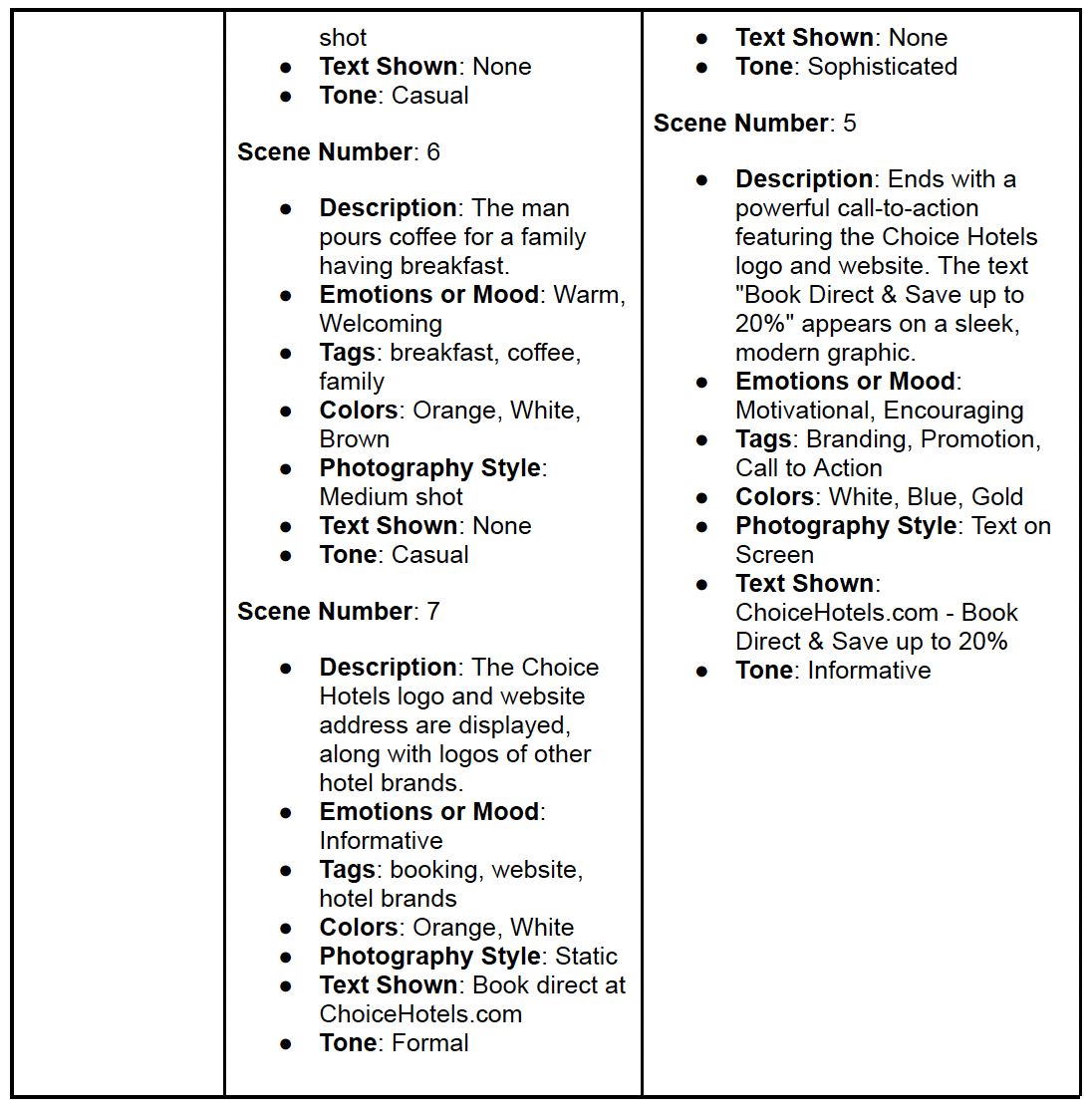}

\end{figure}

\end{document}